\documentclass[conference]{IEEEtran}
\IEEEoverridecommandlockouts
\usepackage{cite}
\usepackage{amsmath,amssymb,amsfonts}
\usepackage{graphicx}
\usepackage{textcomp}
\usepackage[whole]{bxcjkjatype} % for Japanese
\usepackage{multirow}
\usepackage{subcaption}
\usepackage[table,xcdraw]{xcolor}
\usepackage{threeparttable}
\usepackage{eucal}
\usepackage{balance}
\usepackage{mathrsfs}  
\usepackage{comment}
\usepackage{algorithm}
\usepackage[noend]{algpseudocode}
\usepackage{float}
\usepackage{textcomp}
\usepackage{mathcomp}
\usepackage{bm}
\usepackage{cite}
\usepackage{url}
\usepackage{flushend} % 最終ページの下側調整

\newcommand{\myvec}[1]{\mathbf{#1}}

\newcommand{\myset}[1]{\mathit{#1}}

\def\BibTeX{{\rm B\kern-.05em{\sc i\kern-.025em b}\kern-.08em
    T\kern-.1667em\lower.7ex\hbox{E}\kern-.125emX}}
\begin{document}

\title{MPC Builder for Autonomous Drive:\\
Automatic Generation of MPCs for Motion Planning and Control\\
% {\footnotesize \textsuperscript{*}Note: Sub-titles are not captured in Xplore and
% should not be used}
\thanks{This work was supported by J-QuAD DYNAMICS Inc., Tokyo, Japan. This work was also supported by a research grant (C) from Tateisi Science and Technology Foundation.
}
}

\author{\IEEEauthorblockN{1\textsuperscript{st} Kohei Honda}
\IEEEauthorblockA{\textit{Department of Mechanical Systems Engineering} \\
\textit{Nagoya University}\\
Nagoya, Japan \\
honda.kohei.b0@s.mail.nagoya-u.ac.jp}
\and
\IEEEauthorblockN{2\textsuperscript{nd} Hiroyuki Okuda}
\IEEEauthorblockA{\textit{Department of Mechanical Systems Engineering} \\
\textit{Nagoya University}\\
Nagoya, Japan \\
h\_okuda@nuem.nagoya-u.ac.jp}
\and
\IEEEauthorblockN{3\textsuperscript{rd} Tatsuya Suzuki}
\IEEEauthorblockA{\textit{Department of Mechanical Systems Engineering} \\
\textit{Nagoya University}\\
Nagoya, Japan \\
t\_suzuki@nuem.nagoya-u.ac.jp}
\and
\IEEEauthorblockN{4\textsuperscript{th} Akira Ito}
\IEEEauthorblockA{\textit{Department of Mechanical Systems Engineering} \\
\textit{Nagoya University}\\
Nagoya, Japan \\
akira.ito@mae.nagoya-u.ac.jp}
% \and
% \IEEEauthorblockN{5\textsuperscript{th} Given Name Surname}
% \IEEEauthorblockA{\textit{dept. name of organization (of Aff.)} \\
% \textit{name of organization (of Aff.)}\\
% City, Country \\
% email address or ORCID}
% \and
% \IEEEauthorblockN{6\textsuperscript{th} Given Name Surname}
% \IEEEauthorblockA{\textit{dept. name of organization (of Aff.)} \\
% \textit{name of organization (of Aff.)}\\
% City, Country \\
% email address or ORCID}
}

\maketitle

\begin{abstract}
This study presents a new framework for vehicle motion planning and control based on the automatic generation of model predictive controllers (MPCs) named MPC Builder. In this framework, several components necessary for MPC, such as prediction models, constraints, and cost functions, are prepared in advance. The MPC Builder then generates various MPCs online in a unified manner according to traffic situations.
This scheme enabled us to represent various driving tasks with less design effort than typical switched MPC systems. 
The proposed framework was implemented considering the continuation/generalized minimum residual (C/GMRES) method optimization solver, which can reduce computational costs. Finally, numerical experiments on multiple driving scenarios were presented.

\end{abstract}


\section{INTRODUCTION}
\label{sec:introduction}

% \begin{itemize}
%     \item どういう問題を解きたいのか: 様々な環境やタスクに統一的に対応できるMPC
%     \item なぜその問題を解くべきなのか: MPCを実環境で使うために
%     \item その問題を（既存の手法で）解くことは、なぜ難しいのか: 1つのMPCで全ての環境を解くのは難しい，MPC自体をモードごと切り替える方法はたくさんのMPCを予め設計しておかないと多様な運転行動が実現できない
%     \item 上記の困難さを解決する、キーとなるアイデアはなにか。こういうアプローチ・コンセプトでいきます、というシングルセンテンスの主張: MPCを要素分解して，実時間で組み合わせながら走行する
%     \item なぜ、上記のアイデアは、困難さを解決するのか？: MPCを直に切り替えるよりも，サブシステムの再利用性が高まるので，少ない要素数で多様な運転を表現できる
%     \item 上記のアイデアは、提案手法の中で、どのように実現・実装されるか: (1) primitive同士のオペレータを定義 (2) 安定性を上げるために種類分類 (3) 状態空間が大きくなっても実時間になるようにcgmresで解く 
%     \item 提案手法は、どのように評価されるのか、評価基準はなにか？: 4つのシナリオで成功率を見た
% \end{itemize}

Autonomous driving (AD) is expected to reduce traffic accidents and improve transportation comfort. Motion planning is an essential component for realizing AD. The motion planner controls the ego vehicle to accomplish various driving tasks, such as following the leading vehicle, changing lanes, overtaking, and pausing, while considering traffic rules, passenger comfort, and safety.

Model predictive control (MPC) \cite{diehl2011numerical} is a promising vehicle motion planning and control technique. 
As MPC is a finite-time optimal control based on a receding horizon scheme, flexible motion planning and control can be achieved by incorporating various control requirements into cost functions and constraints. MPC performs well particularly in dynamic environments by embedding the prediction models of surrounding road users into the constraints of the optimization problem \cite{interactioin_aware_MPC}.
Consequently, MPC has been applied to complex driving scenarios in the presence of other road users, such as lane changing \cite{hang2020integrated, nilsson2015longitudinal, MPC_lc_sugie}, obstacle avoidance \cite{MPC_obs_avoid, guo2018simultaneous}, pedestrian avoidance \cite{MPC_ped_avoid, MPC_ped_avoid_tran}, adaptive cruise control (ACC) \cite{MPC_ACC}, and turning and crossing at intersections \cite{batkovic2019real, riegger2016centralized}.

\begin{figure}[t]
  \centering
  \includegraphics[width=1.0\linewidth]{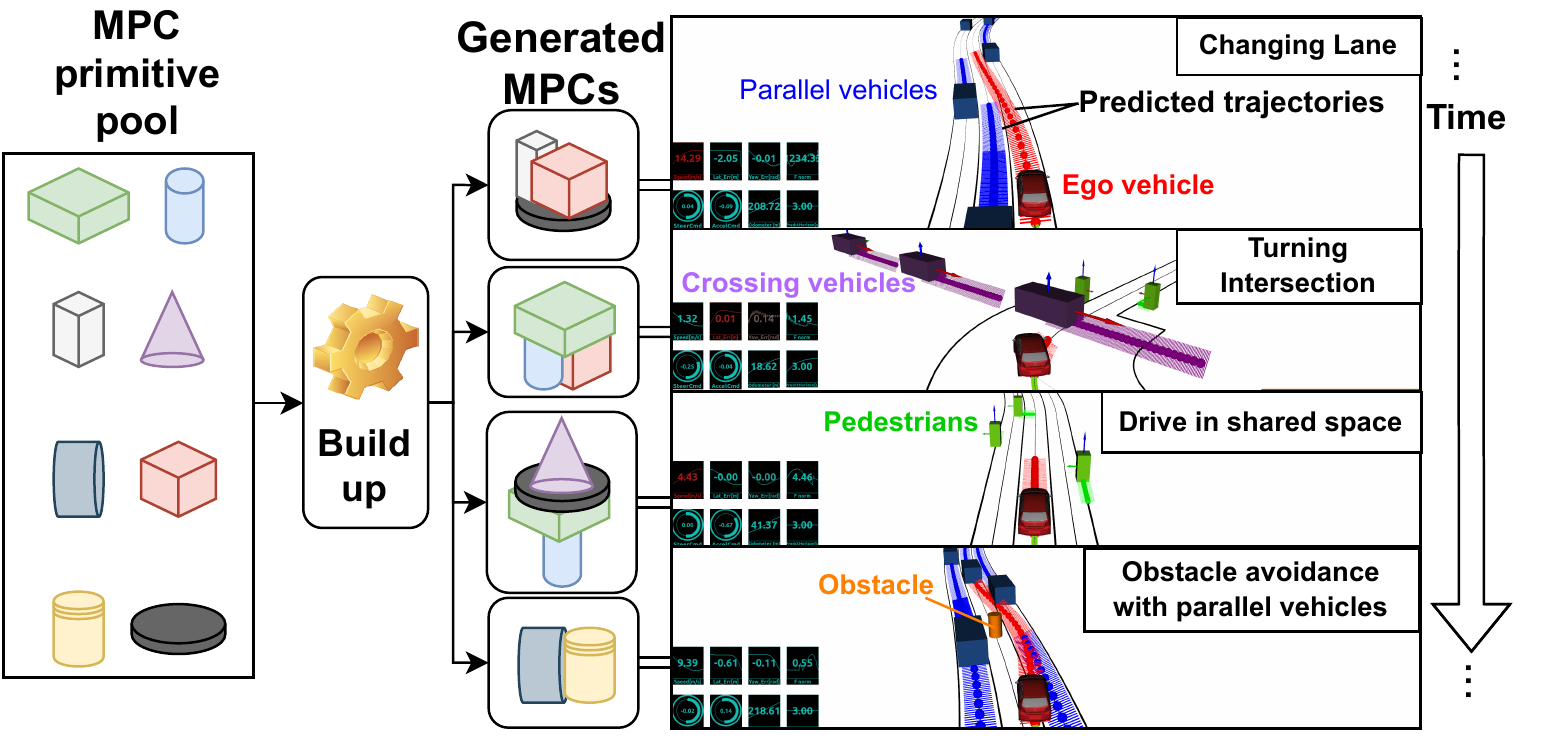}
	\caption{We propose a motion planning and control framework for the automatic generation of MPC to represent various driving tasks. Our proposed framework defines the primitives in advance and builds them online to generate various MPCs according to the traffic situation. The results can be found at https://youtu.be/15J2p26OoLI} 
 % \vspace{-3mm}
  \label{fig:crown_jewel}
\end{figure}

However, to apply MPC to an actual traffic environment, addressing various driving tasks and preparing many MPCs in the design stage are required. 
% \honda{この上の文が強すぎる? contraintの合成とかでもいいし}
This requires intensive design efforts for the AD designers. 
Thus, developing a unified scheme for MPCs that can deal with diverse environments and multiple driving tasks is recommended. 
Previous studies have proposed switched MPC systems that switch multiple MPCs designed for specific driving tasks \cite{zhang2019novel, zhang2020surrounding, vom2021safe, wang2018predictive}. 
% Although these methods can represent multiple driving tasks, the number of available situations remains limited because the number of prepared MPCs is sufficiently large to address many driving tasks.
While these methods are capable of handling various driving tasks, the number of scenarios they can address is limited due to the insufficient number of MPCs available to cover all possible driving situations.

Based on this information, this study presents a new framework for the automatic generation of MPCs, called MPC Builder, which online designs MPCs for various driving tasks according to traffic situations. 
The proposed framework first decomposes the general formulations of MPC into reusable primitives as representations of subtasks. 
The primitives are defined as a set of state spaces, prediction models, cost functions, and constraints to achieve a control requirement. 
The proposed framework then combines the primitives using a binary operator for the primitives to generate optimization problems in real-time, as shown in Fig. \ref{fig:crown_jewel}.
This scheme can represent diverse driving tasks sequentially and in parallel with fewer design elements than the switched MPC system, because the primitives can be replaced and add-on as common components of multiple MPCs.

The generated MPCs can vary the state space and the prediction models depending on the other road users considered.
In order to work in real-time, even in high-dimensional state spaces, we implemented the proposed system considering the continuation/generalized minimum residual (C/GMRES) method \cite{CGMRES}, which is a fast nonlinear MPC solver based on the continuity of the optimal solution. 
Through numerical simulations, four typical driving scenarios were targeted, and the driving behaviors and real-time performance of the proposed method were demonstrated.

% The key contributions of this study are summarized as follows: 
% \begin{enumerate}
%     \item A new framework for the automatic generation of MPCs named MPC Builder is proposed. The concept of the framework is to compose various MPCs from a set of primitives that represent the reusable components of driving tasks.
%     \item The primitives are defined as a set of state spaces, prediction models, cost functions, and constraints to achieve a control requirement. We propose a binary operator to compose the primitives to generate the optimization problems of MPC.
%     \item The proposed framework was demonstrated by numerical simulation for several driving scenarios.
% \end{enumerate}
\section{RELATED WORK}
\label{sec:related_work}

AD requires diversified driving tasks to adapt to complex traffic environments. 
Many studies on AD have addressed single tasks, such as highway cruising and obstacle avoidance. 
Previous studies have addressed multi-task vehicle motion planning in MPC frameworks. Kim et al. applied an adaptive potential field to represent various driving behaviors \cite{kim2018trajectory}. 
Liu et al. proposed a nonlinear MPC that performs multiple driving tasks by convexly relaxing the mixed-integer problem \cite{liu2017path}. 
Although this MPC can implicitly represent various driving tasks as a convex relaxation of the mixed-integer problem, realizing more driving tasks in a single MPC remains challenging. Some studies have proposed switched MPC systems that switch MPCs according to the driving situation \cite{zhang2019novel, zhang2020surrounding, vom2021safe, wang2018predictive}. 
These methods switch weight parameters \cite{MPC_lc_sugie, zhang2019novel}, reference speed and lane \cite{zhang2020surrounding}, safety constraints \cite{vom2021safe}, and cost functions and constraints \cite{wang2018predictive} to accomplish various driving tasks depending on the traffic situation. 
%Therefore, hybrid MPC requires a large implementation space in both hardware and software.
% \okuda{These hybrid MPC approaches have to use common 
% state space and prediction model for the targeting multiple tasks. 
% However, sometime the problem frame, state space in other words, should be 
% extended when the number of agents (surrounding cars and/or pedestrian)
% considered in the problem was changed in realistic driving situation.
% }
Although these MPCs are designed for different driving tasks, they share common elements. As a result, the switched MPC system is redundant for various driving tasks.
The proposed framework reduces the redundancy and implementation effort by representing multiple MPCs by combining common primitives.
The proposed framework is also a general concept of switched MPC system that switches not only the parameters of the cost function and constraints but also the state space and the prediction model. 

The proposed framework represents various MPCs for various driving tasks in a unified manner. It is inspired by the Riemannian Motion Policies (RMP) \cite{RMP}, which combines primitive motion policies for generating a single motion. Some studies have applied RMP to AD \cite{meng2019rmp}. As RMP is based on stationary dynamics, it cannot manage the transient dynamics of the control target, whereas our proposed method can consider any dynamic model by combining the elements of MPC.

% Hierarchical Reinforcement Learning Method for Autonomous Vehicle Behavior Planning∗

% survey: https://arxiv.org/pdf/2105.14218.pdf

% multiple driving task planning

% \honda{下のやつは論文を読んでから問題意識を追記する}
% The approach of synthesizing MPCs is carried out in this study \cite{fan2021controller}.
% 制御器合成という話で，自動で制御器を作るという内容のものなのでちょっと違うかも
% However, it is applied only to linear systems. 
% It is also pointed out that computation complexity increases as the number of state spaces.
\section{MPC BUILDER}

\subsection{Concept of MPC Builder}
\label{sec: concept_proposed_method}

\begin{figure}[t]
  \centering
  \includegraphics[width=1.0\linewidth]{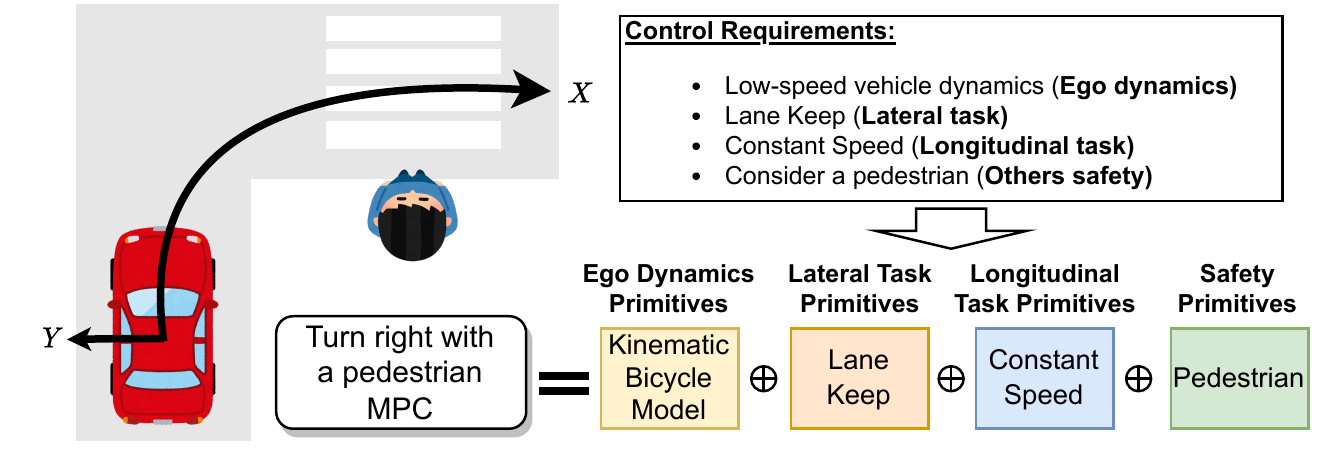}
	\caption{Example of MPC composition scheme. We decompose the control requirements into MPC primitives and combine them to generate MPCs. For example, we represent turn-right with a pedestrian MPC by composing four primitives.}
 % \vspace{-5mm}
  \label{fig:example_composition}
\end{figure}

We propose the MPC Builder, which provides a unified framework for generating MPCs for various AD tasks.
The concept of the MPC Builder is to compose MPCs to achieve various tasks from a set of primitives that express the common elements of the target tasks.
The primitives are small control requirements that are defined in advance.
The requirements include the following.
% \begin{itemize}
%     \item To consider vehicle dynamics
%     \item To consider the safety of other agents
%     \item To achieve driving objectives
% \end{itemize}
\begin{itemize}
    \item Predictive models for the AD vehicle and/or other agents
    \item Safety constraints for other agents and/or obstacles
    \item To achieve a part of the driving objective (subtask)
\end{itemize}
For example, in a right turn with a pedestrian, as shown in Fig. \ref{fig:example_composition}, 
the requirements are to maintain the lane and speed while considering low-speed vehicle dynamics and pedestrian safety. 
% We call the minimum required components of an MPC to achieve and satisfy these requirements \textit{MPC primitive}.
% The MPC Builder concept can represent various driving tasks as a combination by assembling a finite number of pre-pooled MPC primitives.
In the proposed framework, these control requirements are defined in small MPC components called \textit{MPC primitive}, which are combined to represent driving tasks.
Some of these control requirements are not only available for right-turning, but also for other driving tasks.
Therefore, we decompose some MPC for specific driving tasks into MPC primitives and combine them to represent various driving tasks.

\subsection{System Overview}

\begin{figure*}[t]
  \centering
  \includegraphics[width=1.0\linewidth]{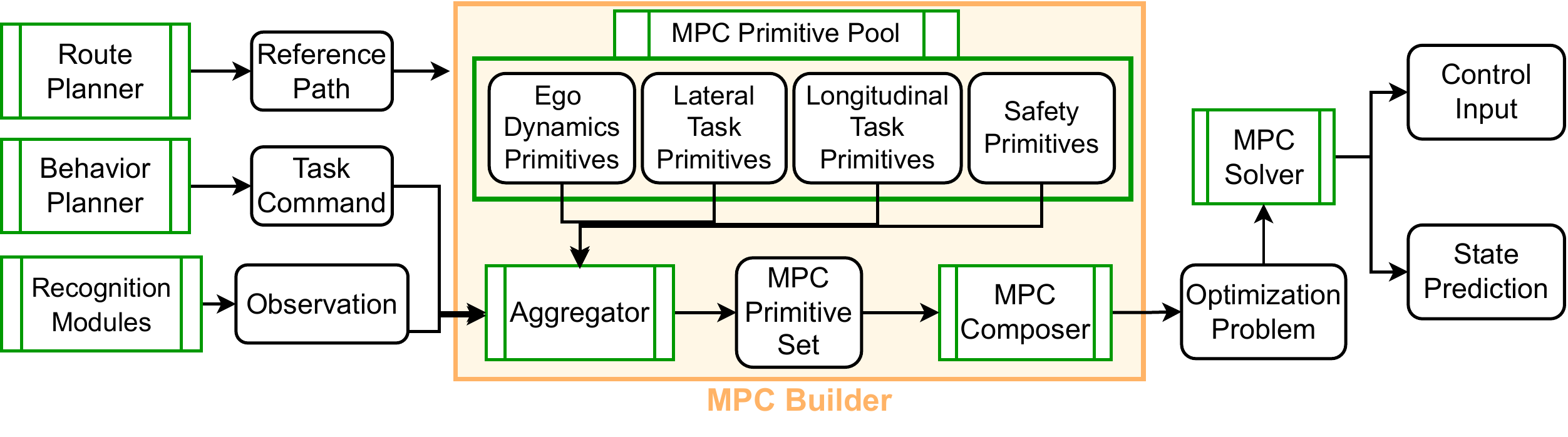}
  % \vspace{-3mm}
	\caption{System overview. MPC Builder online generates the optimization problem at the time by receiving reference paths, driving task commands, and observations from higher-level modules. MPC Builder pools MPC primitives and composes them to generate the optimization problem. Then, the MPC solver optimizes the control input and state prediction in real-time.}
  % \vspace{-5mm}
  \label{fig:system_overview}
\end{figure*}

Based on the aforementioned ideas, 
we propose a system that performs various driving tasks by building MPCs according to the traffic situation.
Figure \ref{fig:system_overview} shows the proposed system outline.
The MPC Builder realizes the desired tasks by receiving reference path information from the route planner and observation from the recognition module with localization and detection.
% Driving behavior is a symbolic task command with parameters to realize the desired behavior (e.g., changing lanes, stopping 100m ahead, etc).
The task command is a symbolic command with parameters to realize the desired behavior (e.g., changing lanes, stopping 100m ahead, etc).
The MPC Builder constructs the optimization problem online to be solved in the MPC framework using this information. 
Finally, the MPC solver solves this optimization problem to calculate and output the control inputs and predictive states in each control cycle.

The MPC Builder consists of three modules: \textit{MPC Primitive Pool}, \textit{Aggregator}, and \textit{MPC Composer}.
\textit{MPC Primitive Pool} stores the MPC primitives defined later (section \ref{sec:mpc_primitive_definition}).
\textit{Aggregator} selects and makes a set of the required MPC primitives according to the driving task command and observation (section \ref{sec:aggregator}).
\textit{MPC Composer} formulates the optimization problem at the control cycle by composing all elements of a set of MPC primitive (section \ref{sec:mpc_composition}).

\subsection{Definition of MPC primitive}
\label{sec:mpc_primitive_definition}
Here, we define the MPC primitive, which is the core idea of the proposed method.

\subsubsection{General formulation of MPC}
\label{subsec: general_formulation_mpc}
This section reviews the general formulation of the optimization problem before defining MPC primitives.
MPC finds a series of optimal control inputs by solving the optimization problem with the finite-time future prediction of every control cycle.
We can obtain the state prediction and optimal control input by solving the following optimization problem at time $t$: 
\begin{align}
  & \textbf{Find: } \hat{\myvec{x}}(k|t) \; \in \myset{X} \in \mathbb{R}^n,\;\; \forall k\in \{1, \dots ,N\},\label{eq:mpc_x}\\
  & \qquad \; \; \; \hat{\myvec{u}}(k|t) \in \myset{U} \in \mathbb{R}^m, \;\; \forall k\in \{0, \dots ,N-1\},\label{eq:mpc_u}\\
  &\textbf{Min.: } J_{0:N}= \sum_{k=0}^{N} J_k ( \hat{\myvec{x}}(k|t), \hat{\myvec{u}}(k|t)),\label{eq:mpc_j}\\
  &\textbf{S.t.: } \hat{\myvec{x}}(0|t) = \myvec{x}(t), \\
  & \qquad \hat{\myvec{x}}(k+1|t) = \hat{\myvec{x}}(k|t) + \myvec{f}(\hat{\myvec{x}}(k|t), \hat{\myvec{u}}(k|t)) \Delta \, \tau,\label{eq:mpc_f}\\
  % Eulerの形式で有る必要はないので変えてみる．\tauってfの積としてくっつく必要あるのか？
  % & \honda{\qquad \hat{\myvec{x}}(k+1|t) = \myvec{f}(\hat{\myvec{x}}(k|t), \hat{\myvec{u}}(k|t)), }\label{eq:mpc_f}\\
  &\qquad \myvec{g}_{0:N}(\hat{\myvec{x}}(k|t), \hat{\myvec{u}}(k|t)) \leq \myvec{0}, \;\; \myvec{h}_{0:N}(\hat{\myvec{x}}(k|t), \hat{\myvec{u}}(k|t)) = \myvec{0}, \label{eq:mpc_h}
\end{align}
where $\myvec{x}(t)$ is the observed value of the state vector $\myvec{x}$ at time $t$; 
$\hat{\;}$ denotes the predicted values and $\hat{\myvec{x}}(k|t)$ and $\hat{\myvec{u}}(k|t)$ are the predicted state and control input vectors for the $k$-th step future at time $t$. $\myset{X}$ and $\myset{U}$ denote the state and control spaces, respectively. 
$N$ and $\Delta \tau$ are the length and time intervals of the prediction horizon, respectively. 
$J_{0:N}: \myset{X} \times \myset{U} \rightarrow \mathbb{R}$ is a cost function that is minimized.
% \honda{ここの射影の引数がおかしい．J\_kならわかる -> いや，これはおかしくない．状態空間は変わらないはず}
% $\myvec{f}: \myset{X} \times \myset{U} \rightarrow \myset{X}$ is a discrete-time state-prediction function that represents the dynamics of the control target.
$\myvec{f}: \myset{X} \times \myset{U} \rightarrow \myset{X}$ is a state-prediction function with Euler discretization that represents the dynamics of the control target.
% \honda{Euler法にする必要は無いが，discrete methodは同一で無いとだめな気がする}
$\myvec{g}_{0:N}: \myset{X} \times \myset{U} \rightarrow \mathbb{R}^{n_g}$ and $\myvec{h}_{0:N}: \myset{X} \times \myset{U} \rightarrow \mathbb{R}^{n_h}$ are function vectors of the inequality and equality constraints, respectively. 
The notation $*_{0:N}$ denotes that the function is applied to the entire prediction horizon from $k=0$ to $N$.
$n_g$ and $n_h$ are the sizes of the inequality and equality constraints, respectively.
The prediction horizon length and control horizon length are the same, $N$ in this study.

In Eqs. (\ref{eq:mpc_x}) to (\ref{eq:mpc_h}), 
the optimization problem that must be set in the MPC solver at time $t$ is identical to the following tuple:
\begin{align}
    & \myset{O}_t  = \{ \myset{U}, \myset{X},  \myvec{f}, J_{0:N}, \myvec{g}_{0:N}, \myvec{h}_{0:N} \}.
    \label{eq:mpc_problem}
\end{align}
Note that the variables $\hat{\myvec{x}}(k|t)$ and $\hat{\myvec{u}}(k|t)$ are instantiated in the solver but are not included in the definition of $\myset{O}_t$.

\subsubsection{Definition of MPC primitive}
\label{subsection:mpc_primitive_definition}

The proposed framework decomposes the optimization problems represented by Eq. (\ref{eq:mpc_problem}) into small components, called MPC primitives, and assembles the optimization problem $\myset{O}_t$. 
With MPC primitives as design elements, we assume that they are designed manually from the control requirements of multiple driving tasks as described in section \ref{sec: concept_proposed_method}.
In Eq. (\ref{eq:mpc_problem}), the smallest primitives of the optimization problem are $\myset{U}, \myset{X},  \myvec{f}, J_{0:N}, \myvec{g}_{0:N}$, and $\myvec{h}_{0:N}$.
However, these smallest primitives alone are not sufficient to represent the control requirements.
For example, a cost function and constraints are required to express the constant speed control requirement, as shown in Fig. \ref{fig:example_composition}.
Thus, we define the MPC primitive $\mathcal{P}$ as a tuple to represent the control requirements:
\begin{align}
    & \mathcal{P} = \{ \myset{X}, \myvec{f}, J_{0:N}, \myvec{g}_{0:N}, \myvec{h}_{0:N} \},
    \label{eq:mpc_primitive_definition}
\end{align}
where each element can be an empty set, different from Eq. (\ref{eq:mpc_problem}). 
$\myset{U}$ is also assumed to be common for all aimed tasks and MPC primitives because the control target vehicle is identical.

\subsection{Composition of MPC primitives}
\label{sec:mpc_composition}

\begin{algorithm}[t]
  \caption{Binary operator $\oplus$ for MPC primitive}
  \label{alg:additive_operator}
  \begin{algorithmic}[1]
  
  \Require{Common control input vector $\myvec{u} \in \myset{U}$}
  \Function{add\_mpc\_primitive}{$\myset{P}_i$, $\myset{P}_j$}
        \State{$\{ \myset{X}_i, \myvec{f}_i, J_{0:N}^i, \myvec{g}_{0:N}^i, \myvec{h}_{0:N}^i \} \gets \myset{P}_i$}
        \State{$\{ \myset{X}_j, \myvec{f}_j, J_{0:N}^j, \myvec{g}_{0:N}^j, \myvec{h}_{0:N}^j \} \gets \myset{P}_j$}
        \State{$\myset{X}_{\rm{ij}} = \myset{X}_{i} \times \myset{X}_{j}$}\label{step:state_space}\\
        \Comment{cartesian product of state space}
        \State{$\myvec{x}_{\rm{ij}} \gets \myset{X}_{\rm{ij}}$}
        \Comment{get ordered state vector}
        \State{$f_{\rm{ij}}(\myvec{x}_{\rm{ij}}, \myvec{u})=f_{i}(\myvec{x}_{\rm{ij}}, \myvec{u}) \times f_{j}(\myvec{x}_{\rm{ij}}, \myvec{u})$}\label{step:prediction_model}\\
        \Comment{extend state prediction function}
        \State{$J_{0:N}^{\rm{ij}}(\myvec{x}_{\rm{ij}}, \myvec{u}) = J_{0:N}^{\rm{i}}(\myvec{x}_{\rm{ij}}, \myvec{u}) + J_{0:N}^{\rm{j}}(\myvec{x}_{\rm{ij}}, \myvec{u})$}\label{step:cost_func}\\
        \Comment{add cost function}
        \State{$\myvec{g}_{0:N}^{\rm{ij}}(\myvec{x}_{\rm{ij}}, \myvec{u}) = \myvec{g}_{0:N}^{i}(\myvec{x}_{\rm{ij}}, \myvec{u}) \times \myvec{g}_{0:N}^{j}(\myvec{x}_{\rm{ij}}, \myvec{u})$}\label{step:ineq_constraint}\\
        \Comment{extend inequality constrants}
        \State{$\myvec{h}_{0:N}^{\rm{ij}}(\myvec{x}_{\rm{ij}}, \myvec{u}) = \myvec{h}_{0:N}^{i}(\myvec{x}_{\rm{ij}}, \myvec{u}) \times \myvec{h}_{0:N}^{j}(\myvec{x}_{\rm{ij}}, \myvec{u})$}\label{step:eq_contraint}\\
        \Comment{extend equality constrants}
        \\
  \Return{$\myset{P}_{\rm{ij}} = \{ \myset{X}_{\rm{ij}}, \myvec{f}_{\rm{ij}}, J_{0:N}^{\rm{ij}}, \myvec{g}_{0:N}^{\rm{ij}}, \myvec{h}_{0:N}^{\rm{ij}} \}$}
  \EndFunction
  \end{algorithmic}
\end{algorithm}

\textit{MPC Composer} shown in the Fig. \ref{fig:system_overview} composes the optimization problem $\myset{O}_t$ from a set of MPC primitives $\mathcal{P}$.
First, we assume that all MPC primitives are formulated with a common prediction horizon; 
horizon length $N$ and time interval $\Delta \tau$ are common.
Under this assumption, the following additive binary operator $\oplus$ is defined between the two MPC primitives:
% \honda{アルゴリズムに書き換える．indexの所在がわかりやすい}
% \begin{align}
%     &\mathcal{P}_i \oplus \mathcal{P}_j := \{ \myset{X}_{ij}, \myvec{f}_{ij}, J_{0:N}^{ij}, \myvec{g}_{0:N}^{ij}, \myvec{h}_{0:N}^{ij} \},\\
%     &\myset{X}_{ij} = \myset{X}_i \times \myset{X}_j, \quad
%     \myvec{f}_{ij} = \myvec{f}_i \times \myvec{f}_j, \quad
%     J_{0:N}^{ij} = J_{0:N}^{i} + J_{0:N}^{j},\\
%     &\myvec{g}_{0:N}^{ij} = \myvec{g}_{0:N}^{i} \times \myvec{g}_{0:N}^{j}, \quad
%     \myvec{h}_{0:N}^{ij} = \myvec{h}_{0:N}^{i} \times \myvec{h}_{0:N}^{j},
% \end{align}
% The Cartesian product of the state spaces denotes the variable space concatenation and
% $\myvec{f}_{ij}: \myset{X}_{ij} \times U \rightarrow \myset{X}_{ij}$, $J_{0:N}^{ij}: \myset{X}_{ij} \times \myset{U} \rightarrow \mathbb{R}$, $\myvec{g}_{0:N}^{ij}: \myset{X}_{ij} \times \myset{U} \rightarrow \mathbb{R}^{n_g^i + n_g^j}$, and $\myvec{h}_{0:N}^{ij}: \myset{X} \times \myset{U} \rightarrow \mathbb{R}^{n_h^i + n_h^j}$ are the extended state prediction function, the cost function, the inequality constraints, and the equality constraints, respectively. 
\begin{align}
    &\mathcal{P}_i \oplus \mathcal{P}_j := \{ \myset{X}_{ij}, \myvec{f}_{ij}, J_{0:N}^{ij}, \myvec{g}_{0:N}^{ij}, \myvec{h}_{0:N}^{ij} \}, 
\end{align}
where $\oplus$ is the binary operator combining two MPC primitives defined in Algorithm \ref{alg:additive_operator}.
In the Algorithm \ref{alg:additive_operator}, the state space is extended in line \ref{step:state_space}.
The Cartesian product of the state spaces denotes the variable space concatenation.
In lines \ref{step:prediction_model} to \ref{step:eq_contraint}, the state prediction function, cost function, and the constraints are extended in the extended state space: $\myvec{f}_{ij}: \myset{X}_{ij} \times U \rightarrow \myset{X}_{ij}$, $J_{0:N}^{ij}: \myset{X}_{ij} \times \myset{U} \rightarrow \mathbb{R}$, $\myvec{g}_{0:N}^{ij}: \myset{X}_{ij} \times \myset{U} \rightarrow \mathbb{R}^{n_g^i + n_g^j}$, and $\myvec{h}_{0:N}^{ij}: \myset{X} \times \myset{U} \rightarrow \mathbb{R}^{n_h^i + n_h^j}$.
Note that the product with a null vector of $\myvec{v}$ 
is defined as $\myvec{v} \times \emptyset = \myvec{v}$ for vector $\myvec{v} \in \{\myvec{f},\myvec{g},\myvec{h}\}$.
% The additive operator generates a new MPC primitive by combining the two MPC primitives.
Using this operator iteratively, the \textit{MPC Composer} shown in Fig. \ref{fig:system_overview} composes the optimization problem $\myset{O}_t$ from a MPC primitive set $\myset{P}=\{\mathcal{P}_1, \dots \mathcal{P}_M\}$ at every control cycle:
\begin{align}
    \myset{O}_t & = \{U, \; \sum_{i=1}^{M} \mathcal{P}_i \} = \{U, \; \left( \left( \mathcal{P}_1 \oplus \mathcal{P}_2 \right) \oplus \mathcal{P}_3 \right)  \oplus \cdots \}.
    \label{eq:solved_problem}
\end{align}
% The control input and state prediction are obtained by solving the optimization problem $\myset{O}_t$ at time $t$.

\subsection{Example of MPC Composition}
\label{sec:example}

This section illustrates an example of the composition of MPC primitives using the turn-right scenario with a pedestrian, as shown in Fig. \ref{fig:example_composition} introduced in section \ref{sec: concept_proposed_method}.
We prepared the following four primitives: 
kinematic bicycle model (KBM) \cite{vehicle_model_Rajamani}, lane keep, constant speed, and pedestrian primitives to represent this driving scenario.
Table \ref{tab:primitive_formula} lists the detailed formulations of the MPC primitives.
\textit{MPC Composer} composes these primitives and obtains the following optimization problem: 
\begin{align}
  & \textbf{Find: } \hat{\myvec{x}}(k|t) \; \in \{ \myset{X}_{\rm{ego}} \times \myset{X}_{\rm{ped}} \},\;\; \forall k\in \{1, \dots ,N\},\\
  & \qquad \; \; \; \hat{\myvec{u}}(k|t) \in \myset{U}, \;\; \forall k\in \{0, \dots ,N-1\},\\
  &\textbf{Min.: } J_{0:N}= J_{\rm{lk}} + J_{\rm{cs}},\\
  &\textbf{S.t.: } \hat{\myvec{x}}(0|t) = \myvec{x}(t), \\
  &\qquad \hat{\myvec{x}}(k+1|t) = \hat{\myvec{x}}(k|t) + [\myvec{f}_{\rm{kbm}}, \myvec{f}_{\rm{ped}}]^T \Delta \, \tau,\\ 
  &\qquad [\myvec{g}_{\rm{lk}}, \myvec{g}_{\rm{cs}}, \myvec{g}_{\rm{ped}}]^T \leq \myvec{0},
\end{align}
where $\myset{X}_{\rm{ego}}$ and $\myset{X}_{\rm{ped}}$ are the state spaces of ego vehicles and pedestrians, respectively.
The control input space $U = \{\dot{\delta}, \dot{a}\}$ are the time derivatives of the steering angle and acceleration of the ego vehicle, respectively.
$*_{\rm{kbm}}, *_{\rm{lk}}, *_{\rm{cs}}$, and $*_{\rm{ped}}$ are the elements of the kinematic bicycle model, lane keep, constant speed, and pedestrian primitives, respectively.
The control input and state prediction of the ego vehicle and pedestrian can be obtained by solving the optimization problem.
This example shows that the MPC Builder can express a driving task based on the composition of MPC primitives.

\subsection{Classification of MPC primitives}

\begin{figure}[t]
  \centering
  \includegraphics[width=0.9\linewidth]{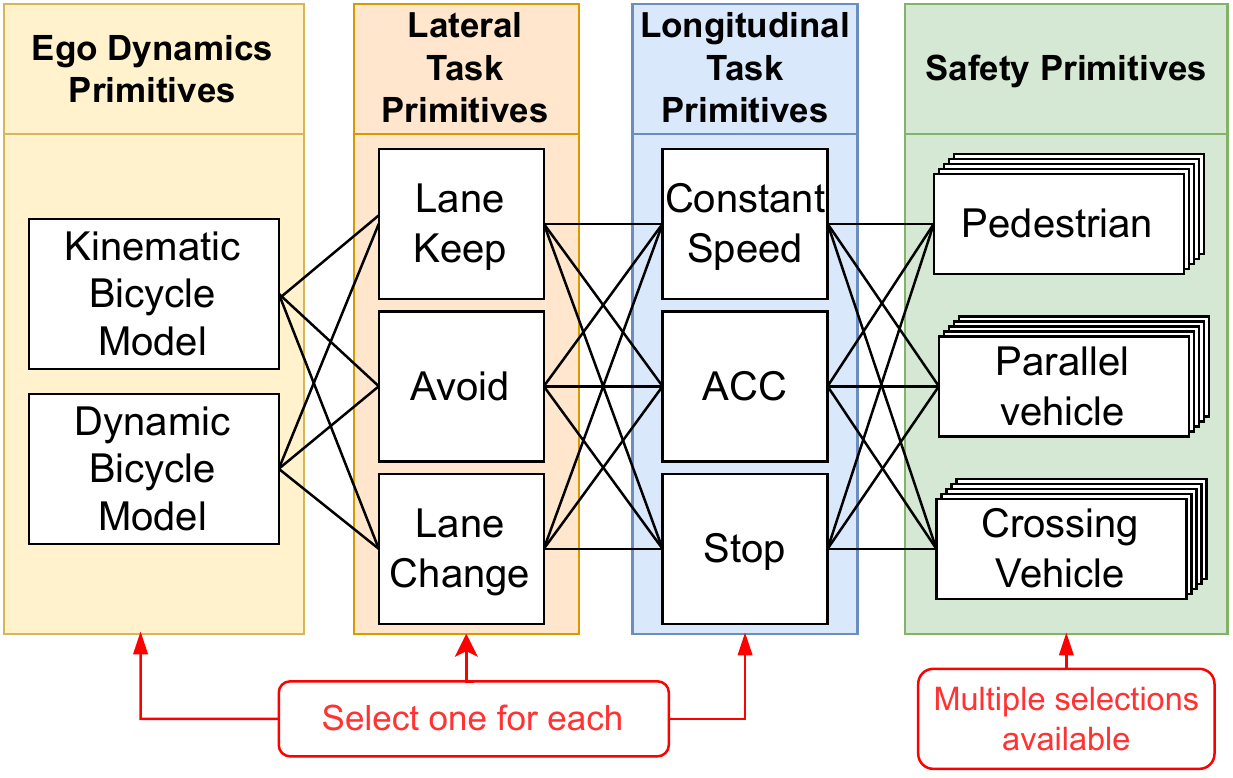}
  % \vspace{-2mm}
	\caption{Classification of MPC Primitives. We classify MPC primitives into four types to consider compatibility and improve reusability. }
  % \vspace{-3mm}
  \label{fig:mpc_primitive_types}
\end{figure}
% \honda{table Iの重みは？本当はどこかのweb pageで公開するといいが，おそらくIVならそこは突っ込まれないだろう}
% \honda{このreasonableがあいまい．要は，意味のある最適化問題が得られるように，ということを言いたい}
MPC primitives should be composed by considering compatibility for obtaining a reasonable optimization problem.
We classified MPC primitives into four types:
ego dynamics, lateral tasks, longitudinal tasks, and safety primitives, as shown in Table \ref{tab:primitive_formula}.
% regarding the physical meanings of the primitives, as shown in Table \ref{tab:primitive_formula}.
Ego dynamics primitives contain the vehicle state and its prediction model with integrated lateral and longitudinal states.
Lateral and longitudinal task primitives are related to the driving objectives expressed by the cost functions and constraints in the Frenet-Serret coordinate \cite{frenet_serret, kloock2019distributed}. 
For example, lane-keep and constant-speed primitives are considered in the turn-right scenario of Fig. \ref{fig:example_composition}.
Safety primitives have state space, state prediction model, cost function, and constraints for predicting the motion of other road users and maintaining safety.

This classification makes it easier to select the appropriate primitives to compose the desired task.
The MPC solver may fail to compute an optimal solution if the selected MPC primitives are contradictory.
Then, only one among ego dynamics, lateral, and longitudinal task primitives can be selected in each control cycle to compose various driving tasks while maintaining consistency.

On the other hand, multiple safety primitives can be activated depending on the number of surrounding pedestrians and vehicles. 
% This realizes a large flexibility in dealing with many combinatorial environmental variations by expressing additive safety primitives.
This provides great flexibility in dealing with many combinatorial environmental variations by expressing additive safety primitives.

\begin{table*}[t]
\centering
\caption{Formulations of MPC primitive}
% \vspace{-2mm}
\label{tab:primitive_formula}
\begin{threeparttable}
\begingroup
\renewcommand{\arraystretch}{1.3}
\begin{tabular}{c|c|ccccc}
\hline
\rowcolor[HTML]{C0C0C0} 
  Type &
  name &
  $\myset{X}$ &
  $J_{0:N}$ &
  $\myvec{f}$ &
  $\myvec{g}_{0:N}$ &
  $\myvec{h}$ \\ \hline
\cellcolor[HTML]{FFFFC7} &
  \cellcolor[HTML]{FFFFC7}KBM &
  $X_{\rm{ego}}$ &
  0 &
  $\myvec{f}_{\rm{KBM}}$ &
  $\emptyset$ &
  $\emptyset$ \\ \cline{2-7} 
\multirow{-2}{*}{\cellcolor[HTML]{FFFFC7}\begin{tabular}[c]{@{}c@{}}Ego\\Dynamics\end{tabular}} &
  \cellcolor[HTML]{FFFFC7}DBM &
  $X_{\rm{ego}}$ &
  0 &
  $\myvec{f}_{\rm{DBM}}$ &
  $\emptyset$ &
  $\emptyset$ \\ \hline
\cellcolor[HTML]{FFCE93} &
  \cellcolor[HTML]{FFCE93}LK &
  $\emptyset$ &
  $\{\|y\|^2+\|\theta\|^2+\|\dot{\theta}\|^2+\|\delta\|^2+\|\dot{\delta}\|^2\}_{Q_{\rm{lk}}}$ &
  $\emptyset$ &
  $ [y_{\rm{min}} - y, y - y_{\rm{max}}, \delta_{\rm{min}} - \delta,  \delta - \delta_{\rm{max}} ]$ &
  $\emptyset$ \\ \cline{2-7} 
\cellcolor[HTML]{FFCE93} &
  \cellcolor[HTML]{FFCE93}Avoid &
  $\emptyset$ &
  $\{\|\theta\|^2+\|\dot{\theta}\|^2+\|\delta\|^2+\|\dot{\delta}\|^2\}_{Q_{\rm{av}}}$ &
  $\emptyset$ &
  \begin{tabular}[c]{@{}c@{}}$ [y_{\rm{min}} - y,  y -  y_{\rm{max}}, \delta_{\rm{min}} - \delta,  \delta - \delta_{\rm{max}},$\\ $d_{\rm{safe}}^{\rm{av}} - \sqrt{(x-x_o)^2+(y-y_o)^2} ]$\end{tabular} &
  $\emptyset$ \\ \cline{2-7} 
\multirow{-3}{*}{\cellcolor[HTML]{FFCE93}\begin{tabular}[c]{@{}c@{}}Lateral\\Task\end{tabular}} &
  \cellcolor[HTML]{FFCE93} LC &
  $\emptyset$ &
  $\{\|y-y_{\rm{ref}}\|^2+\|\theta\|^2+\|\dot{\theta}\|^2+\|\delta\|^2+\|\dot{\delta}\|^2\}_{Q_{\rm{lc}}}$ &
  $\emptyset$ &
  \begin{tabular}[c]{@{}c@{}}$[y_{\rm{min}} - y, y - y_{\rm{max}}, \delta_{\rm{min}} - \delta, \delta - \delta_{\rm{max}},$\\ $d_{\rm{safe}}^{\rm{lc}} - \|x-x_{\rm{pv}}\|]$\end{tabular} &
  $\emptyset$ \\ \hline
\cellcolor[HTML]{96FFFB} &
  \cellcolor[HTML]{96FFFB} CS &
  $\emptyset$ &
  $\{\|v-v_{\rm{ref}}\|^2+\|a\|^2+\|\dot{a}\|^2\}_{Q_{\rm{cs}}}$ &
  $\emptyset$ &
  $[ a_{\rm{min}} - a, a - a_{\rm{max}} ]$ &
  $\emptyset$ \\ \cline{2-7} 
\cellcolor[HTML]{96FFFB} &
  \cellcolor[HTML]{96FFFB}ACC &
  $\emptyset$ &
  $\{\|x-x_{\rm{pv}}-d_{\rm{acc}}\|^2+\|a\|^2+\|\dot{a}\|^2\}_{Q_{\rm{acc}}}$ &
  $\emptyset$ &
  $[a_{\rm{min}} - a, a - a_{\rm{max}}, d_{\rm{safe}}^{\rm{acc}} - \|x-x_{\rm{pv}}\|]$ &
  $\emptyset$ \\ \cline{2-7} 
\multirow{-3}{*}{\cellcolor[HTML]{96FFFB}\begin{tabular}[c]{@{}c@{}}Longitudinal\\Task\end{tabular}} &
  \cellcolor[HTML]{96FFFB}Stop &
  $\emptyset$ &
  $\{\|x-x_{\rm{stop}}\|^2+\|v\|^2+\|a\|^2+\|\dot{a}\|^2\}_{Q_{\rm{stop}}}$ &
  $\emptyset$ &
  $[ a_{\rm{min}} - a, a - a_{\rm{max}}, d_{\rm{safe}}^{\rm{stop}} - \|x-x_{\rm{stop}}\| ]$ &
  $\emptyset$ \\ \hline
\cellcolor[HTML]{9AFF99} &
  \cellcolor[HTML]{9AFF99}PED &
  $X_{\rm{ped}}$ &
  $\{\|v\|^2\}_{Q_{\rm{ped}}}$ &
  $\myvec{f}_{\rm{ped}}$  &
  $[ d_{\rm{safe}}^{\rm{ped}} - \sqrt{(x-x_{\rm{ped}})^2+(y-y_{\rm{ped}})^2} ]$ &
  $\emptyset$ \\ \cline{2-7} 
\cellcolor[HTML]{9AFF99} &
  \cellcolor[HTML]{9AFF99}PV &
  $X_{\rm{pv}}$ &
  0 &
  $\myvec{f}_{\rm{pv}}$  &
  $[ d_{\rm{safe}}^{\rm{pv}} - \sqrt{(x-x_{\rm{pv}})^2+(y-y_{\rm{pv}})^2} ]$ &
  $\emptyset$ \\ \cline{2-7} 
\multirow{-3}{*}{\cellcolor[HTML]{9AFF99}\begin{tabular}[c]{@{}c@{}}Safety\end{tabular}} &
  \cellcolor[HTML]{9AFF99}CV &
  $X_{\rm{cv}}$ &
  $\{\|v\|^2\}_{Q_{\rm{cv}}}$ &
  $\myvec{f}_{\rm{cv}}$  &
  $[ d_{\rm{safe}}^{\rm{cv}} - \sqrt{(x-x_{\rm{cv}})^2+(y-y_{\rm{cv}})^2} ]$ &
  $\emptyset$ \\ \hline
\end{tabular}
\endgroup
\begin{tablenotes}[flushleft]
\item[1] KBM and DBM means kinematic \cite{compare_KBM_realcar} and dynamic \cite{vehicle_model_Rajamani} bicycle models. Lane keep (LK), Avoid, and lane change (LC) are in lateral task primitive. Constant speed (CS), adaptive-cruise-control (ACC), and stop are in longitudinal task primitive. Pedestrian (PED), parallel vehicle (PV), and crossing vehicle (PV) are safety primitive.
\item[2] $\myset{X}_{\rm{ego}}$ is a state space of ego vehicle; $\myset{X}_{\rm{ego}} = [x, y, \theta, \dot{\theta}, v, a, \delta, \dot{\delta}]$. Here, ($x$, $y$, $\theta$) is an ego vehicle pose in the Frenet coordinate \cite{frenet_serret}. 
$v$, $a$, and $\delta$ are the ego vehicle's speed, acceleration, and tire steer angle. 
$\myset{X}_* \in \{ X_{\rm{ped}}, X_{\rm{pv}}, X_{\rm{cv}} \}$ are state spaces of a pedestrian, parallel running vehicle, crossing vehicle; $\myset{X}_* = [x_*, y_*, v_x^*, v_y^*]$ where $(x_*, y_*)$ and $(v_x^*, v_y^*)$ are other agent position and velocity in Frenet coordinate. $(x_o, y_o)$ is also a static obstacle position.
$\myvec{f}_{*} \in \{ \myvec{f}_{\rm{ped}}, \myvec{f}_{\rm{pv}}, \myvec{f}_{\rm{cv}} \}$ are other agent models. We model the 
motions of others with the constant velocity model.
$v_{\rm{ref}}$ and $y_{\rm{ref}}$ are given the reference speed and the y-coordinate value of the lane center from the Route Planner module, as shown in Fig. \ref{fig:system_overview}.
$d_{\rm{acc}}$ is a target relative distance in ACC.
$x_{\rm{stop}}$ is a target stop line position.
$Q_*$ represents the given weight coefficient vectors. $(*_{\rm{min}}, *_{\rm{max}})$ and $d_{\rm{safe}^*}$ are given min./max. values and safety distances.
\end{tablenotes}
\end{threeparttable}
% \vspace{-3mm}
\end{table*}

\section{VALIDATION OF MPC BUILDER CONCEPT}

\subsection{Simulation Scenarios}
\label{sec:simulation_scenario}
To demonstrate the proposed framework, numerical experiments were conducted for the following four driving scenarios.

\subsubsection{Scenario AL (adaptive-cruise-control and lane-changing)}
As shown in Fig. \ref{fig:acc_lanechange}, blue parallel vehicles are randomly spawned and travel with a constant speed. 
A red ego vehicle drives with the ACC at the beginning and changes lanes without collision after a certain time.

\subsubsection{Scenario OA (obstacle avoidance)}
The red ego vehicle avoids the orange obstacles in the presence of multiple random parallel vehicles, as shown in Fig. \ref{fig:avoid}.

\subsubsection{Scenario TI (turning at intersection without signal)}
The ego vehicle turns to the right at an intersection without a signal, as shown in Fig. \ref{fig:example_composition}. 
Random violet-crossing vehicles and green pedestrians at a constant speed across the road, as shown in Fig. \ref{fig:intersection}.
We do not explicitly determine the timing of the right turn; the MPC implicitly determines it.

\subsubsection{Scenario SD (shared road driving)}
The ego vehicle drives slowly on a shared road, where random pedestrians spawn. 
Finally, the ego vehicle stops at the stop line, as shown in Fig. \ref{fig:shared_space}.

\subsection{Implementation Details}
\label{sec:impl_detail}
We implemented a vehicle navigation system using the proposed framework, as shown in Fig. \ref{fig:system_overview} using C++.

\subsubsection{Applied MPC primitives}
To drive through the scenarios described in section \ref{sec:simulation_scenario}, 
eleven MPC primitives;
Two ego dynamics, three lateral tasks, three longitudinal tasks, and three safety primitives, as shown in Table \ref{tab:primitive_formula} were considered.
Thus, theoretically, $2 \times 3 \times 3 \times N_{\rm{ped}} \times N_{\rm{pv}} \times N_{\rm{cv}}$ MPCs can be generated.
Here, $N_{\rm{ped}}$, $N_{\rm{pv}}$, and $N_{\rm{cv}}$ are the maximum numbers of pedestrians, parallel vehicles, and crossing vehicles, respectively.
All MPC primitives are designed in the Frenet coordinate system using reference driving lines planned by the upper-level route planner.

The prediction horizon length and time interval are $N=300$ and $\Delta \tau = 0.01$s, respectively.
This means that the generated MPCs predict 3 seconds future.
The control input space is $U = \{\dot{\delta}, \dot{a}\}$ including the time derivative of the steering angle and acceleration of the ego vehicle.
The control commands are calculated by adding the input to the observed steering angle and acceleration to obtain smooth behavior.

\subsubsection{Aggregator}
\label{sec:aggregator}
\textit{Aggregator} shown in Fig. \ref{fig:system_overview} selects the necessary MPC primitives based on a simple rule.
\textit{Aggregator} selects the lateral and longitudinal task primitives corresponding to the given driving task commands.
Next, the \textit{Aggregator} selects the appropriate ego dynamics primitive,  
KBM primitive for slow speeds ($\leq$ 5.0 m/s), and the dynamic bicycle model primitive otherwise, 
based on the measured vehicle-speed range \cite{aoki_2021_IV}. 
Finally, \textit{Aggregator} adds multiple safety primitives according to the number of observations surrounding the others. 
For example, five parallel vehicle primitives were selected if five parallel vehicles existed.
To reduce the computational load, the number of other vehicles and pedestrians was limited by looking only at their neighbors below a certain threshold. 

%to eight, three, and three.

\subsubsection{MPC solver}

As MPC primitives include nonlinear terms, MPC Builder generates nonlinear optimization problems.
Thus, the continuation/generalized minimum residual (C/GMRES) \cite{CGMRES} method
was used as the optimization solver for the MPC.
The C/GMRES method is an optimization method for nonlinear MPCs based on continuation, and it can achieve fast computation
despite the long prediction horizon and large state space, including the state prediction of the surrounding others.

\begin{figure}
      \begin{minipage}[t]{\linewidth}
    \centering
    \includegraphics[width=\linewidth]{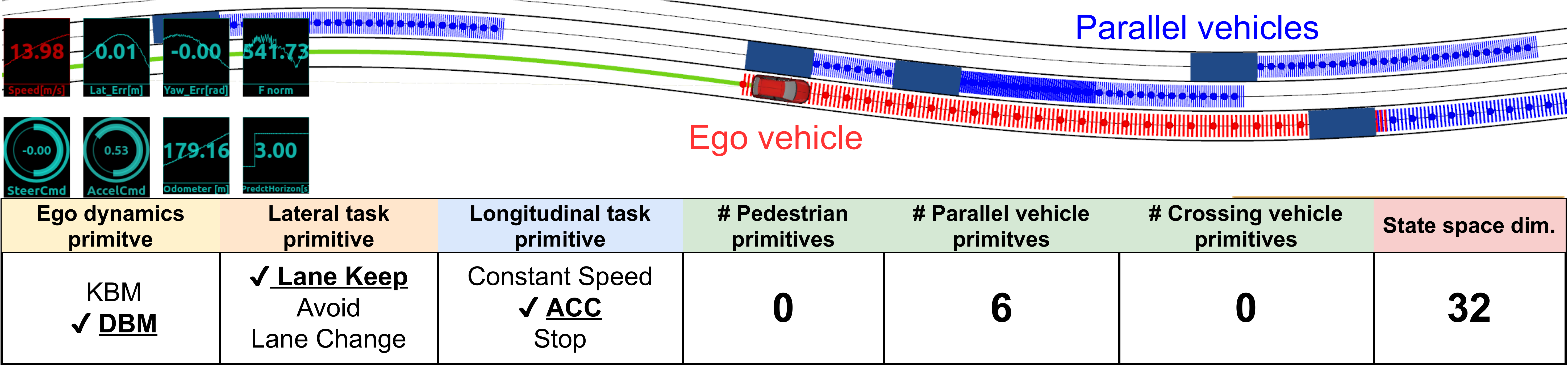}
    \subcaption{Keeping lane with ACC}
    \label{fig:acc_lanechange_1}
  \end{minipage}\\
  \begin{minipage}[t]{\linewidth}
    \centering
    \includegraphics[width=\linewidth]{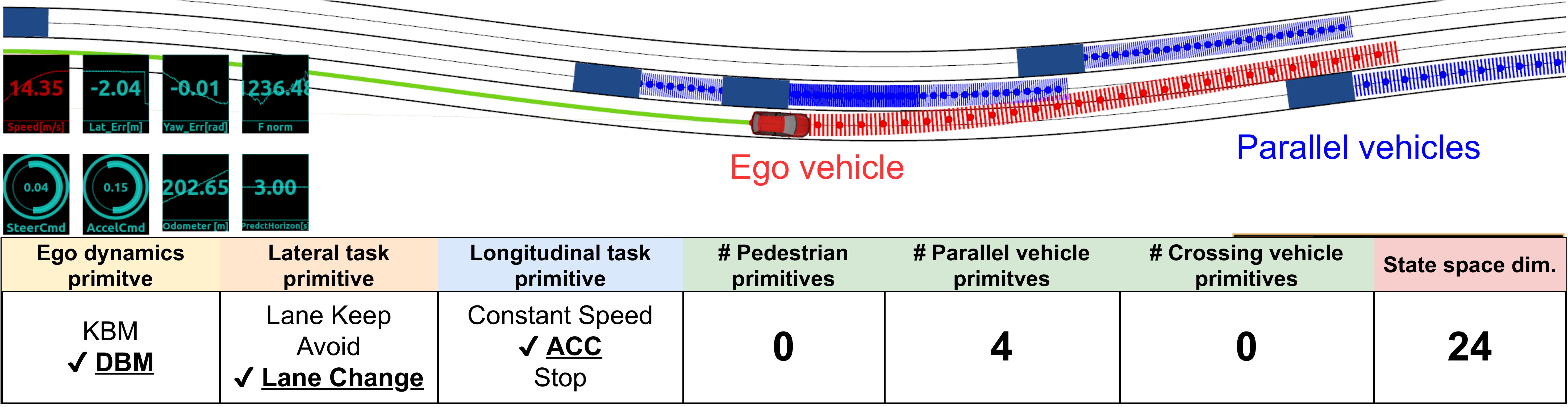}
    \subcaption{Changing lane to the left lane}
    \label{fig:acc_lanechange_2}
  \end{minipage}
  \caption{Result of scenario AL. The ego vehicle followed a front vehicle with lane-keeping (a) and smoothly executed to change lanes without collision (b). We can see that the state space size of the generated MPCs changed in each scene.}
  % \vspace{-5mm}
  \label{fig:acc_lanechange}
\end{figure}

\begin{figure}
      \begin{minipage}[t]{\linewidth}
    \centering
    \includegraphics[width=\linewidth]{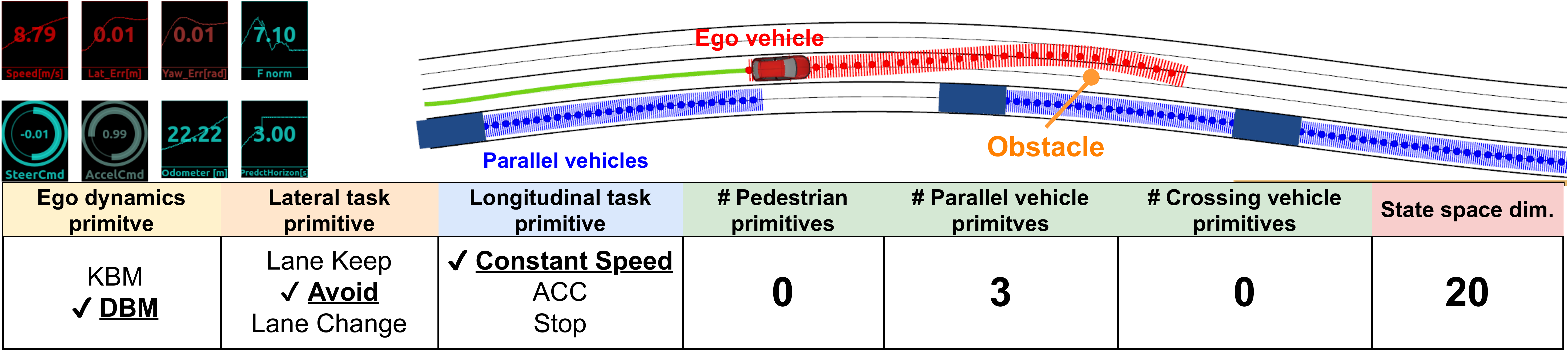}
    \subcaption{Avoiding obstacle to left}
    \label{fig:avoid_1}
  \end{minipage}\\
  \begin{minipage}[t]{\linewidth}
    \centering
    \includegraphics[width=\linewidth]{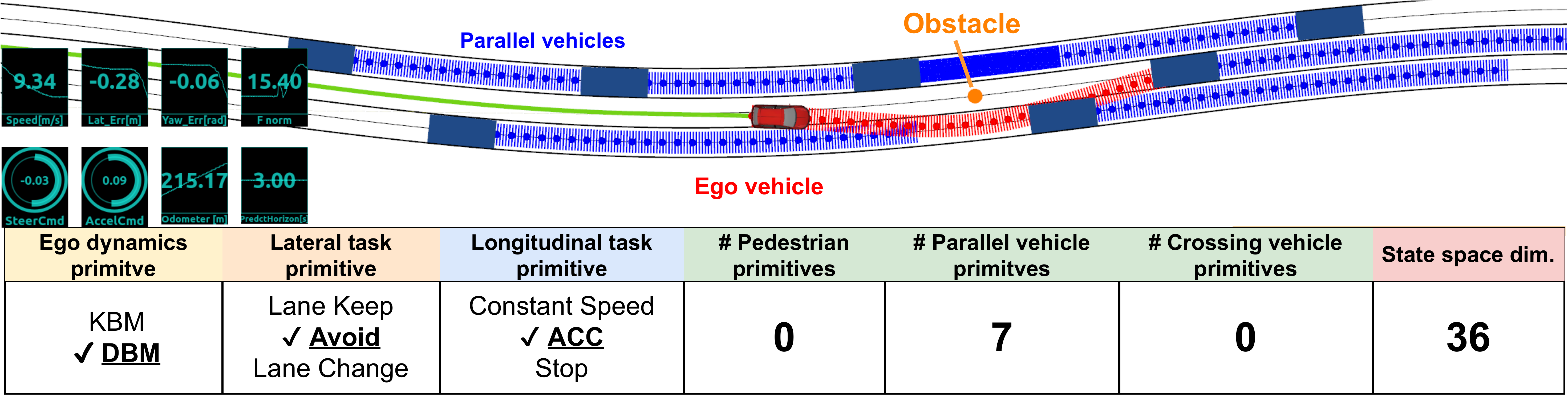}
    \subcaption{Avoiding obstacle to right}
    \label{fig:avoid_2}
  \end{minipage}
  \caption{Result of scenario OA. The ego vehicle successfully avoided an obstacle while surrounded by parallel vehicles. The direction of avoidance is implicitly determined by the generated MPC.}
  % \vspace{-3mm}
  \label{fig:avoid}
\end{figure}

\subsection{Simulation Results}
\label{sec:simulation_result}

In the four driving scenarios (section \ref{sec:simulation_scenario}), we performed 100 drives in each scenario. 
Results for each scenario are shown in Figs. \ref{fig:acc_lanechange} to \ref{fig:shared_space}
\footnote{These results can be found at \url{https://youtu.be/15J2p26OoLI}}.
% \footnote{These results can be found in the attached movie.}.
At the bottom of each figure, we illustrate selected MPC primitives and the state-space size of the generated MPCs at the time instance.
The generated MPC accurately predicts the behavior of the agents and controls the red ego vehicle.
The predicted trajectories of each agent are depicted with dashed lines of the same color as that of the agent.

\subsubsection{Pickup scenes}
Figure \ref{fig:acc_lanechange} shows two successive scenes in AL.
The ego vehicle initially followed the front vehicle in the same lane while maintaining the lane. 
When the lane-change primitive is selected instead of the lane-keep primitive, 
the generated MPC planned trajectory for lane changing without collision.
The state-space sizes of the generated MPC changed with the number of surrounding parallel vehicles.

The two obstacle avoidance scenes for scenario OA are shown in Fig. \ref{fig:avoid}.
The ego vehicle avoids obstacles with the surrounding parallel driving vehicles.
Note that the direction of avoidance was not specified; the generated MPCs implicitly determined the collision-free directions considering the prediction of the surrounding vehicles.

\begin{figure}
      \begin{minipage}[t]{\linewidth}
    \centering
    \includegraphics[width=\linewidth]{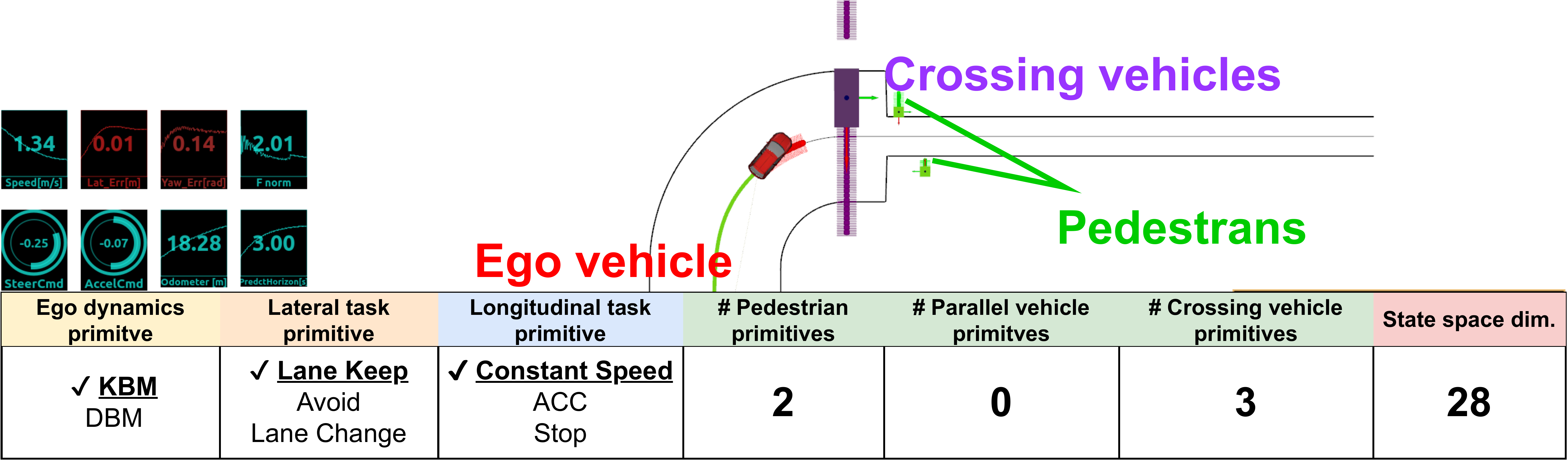}
    \subcaption{Waiting for passing crossing vehicles}
    \label{fig:intersection_1}
  \end{minipage}\\
  \begin{minipage}[t]{\linewidth}
    \centering
    \includegraphics[width=\linewidth]{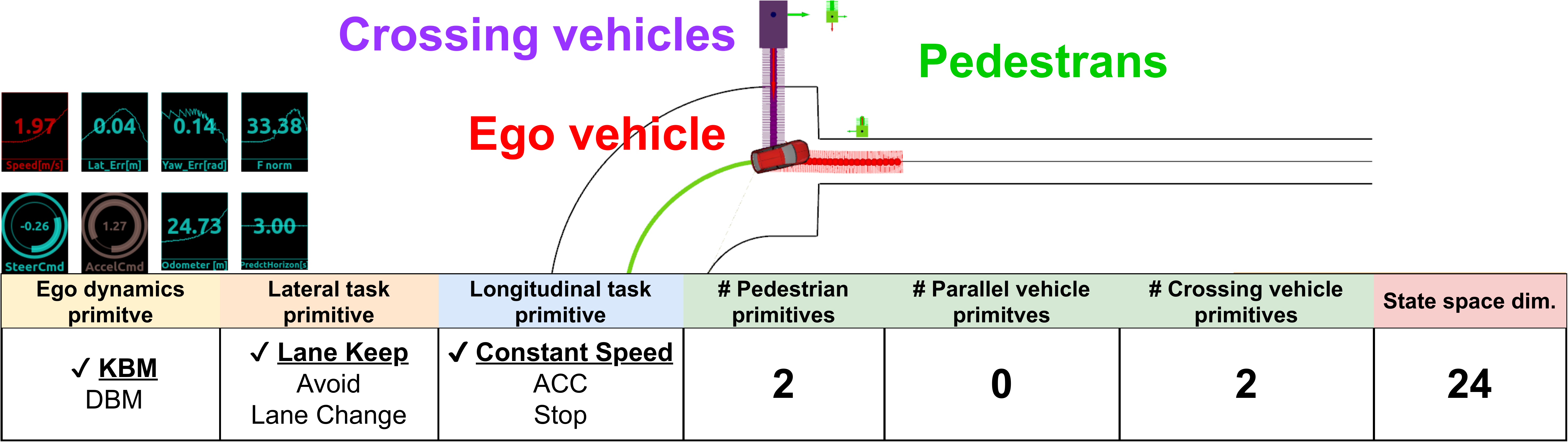}
    \subcaption{Turning right}
    \label{fig:intersection_2}
  \end{minipage}\\
  \begin{minipage}[t]{\linewidth}
    \centering
    \includegraphics[width=\linewidth]{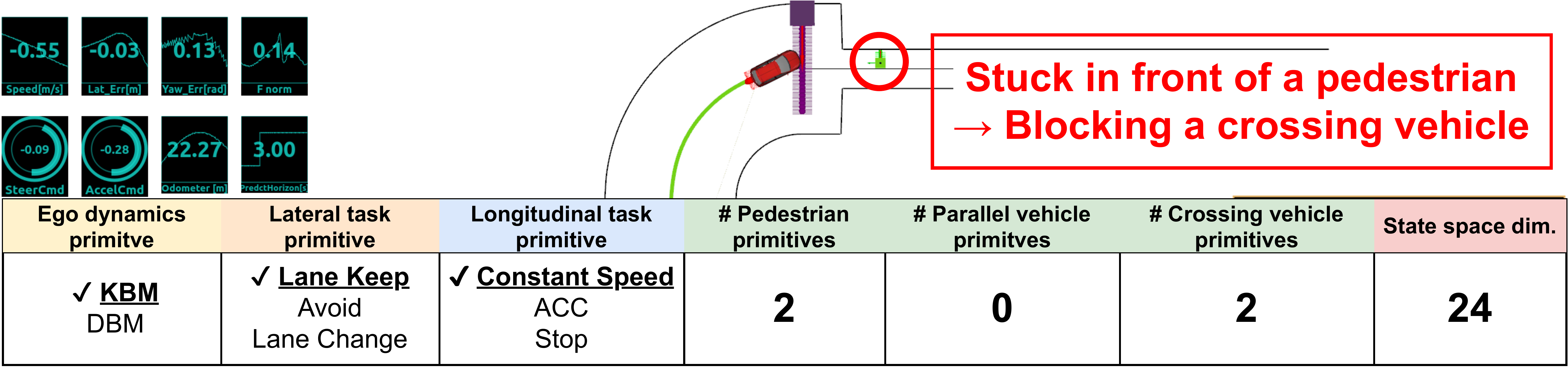}
    \subcaption{Trapping local minima}
    \label{fig:intersection_3}
  \end{minipage}
  \caption{Result of scenario TI. The ego vehicle executed the right turn considering the crossing vehicles and pedestrians. The timing of the right turn was not given and was determined implicitly by MPC. The ego vehicle blocked the crossing vehicle because the generated MPC was trapped in local minima as shown in (c).}
  % \vspace{-2mm}
  \label{fig:intersection}
\end{figure}

Figure \ref{fig:intersection} illustrates the three situations in Scenario TI.
We do not provide the timing of the right turn, and MPC implicitly determines it.
We can see successful cases in (a) and (b) of turning right when considering crossing vehicles and pedestrians.
In Case (c), one pedestrian stopped on the road, 
the ego vehicle started turning and got stuck in front of the pedestrian, while the ego vehicle could stop safely.
In this case, the ego vehicle blocks the crossing vehicle.
This solution was obtained because the generated MPCs were trapped in the local minima.
Although the extension of the prediction horizon solves this issue, it leads to increased computation time. 

\begin{figure}
      \begin{minipage}[t]{\linewidth}
    \centering
    \includegraphics[width=\linewidth]{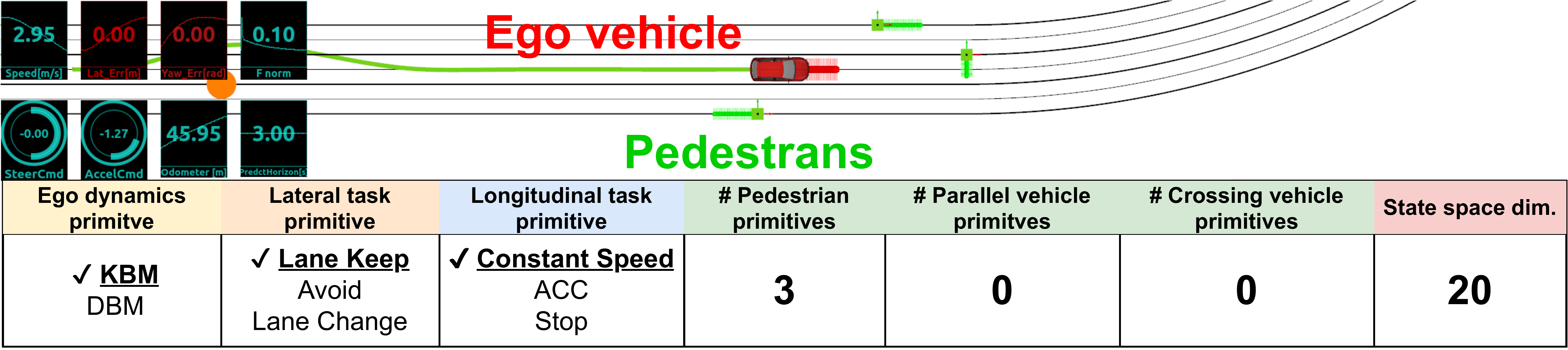}
    \subcaption{Slowing down for pedestrian crossing}
    \label{fig:shared_space_1}
  \end{minipage}\\
  \begin{minipage}[t]{\linewidth}
    \centering
    \includegraphics[width=\linewidth]{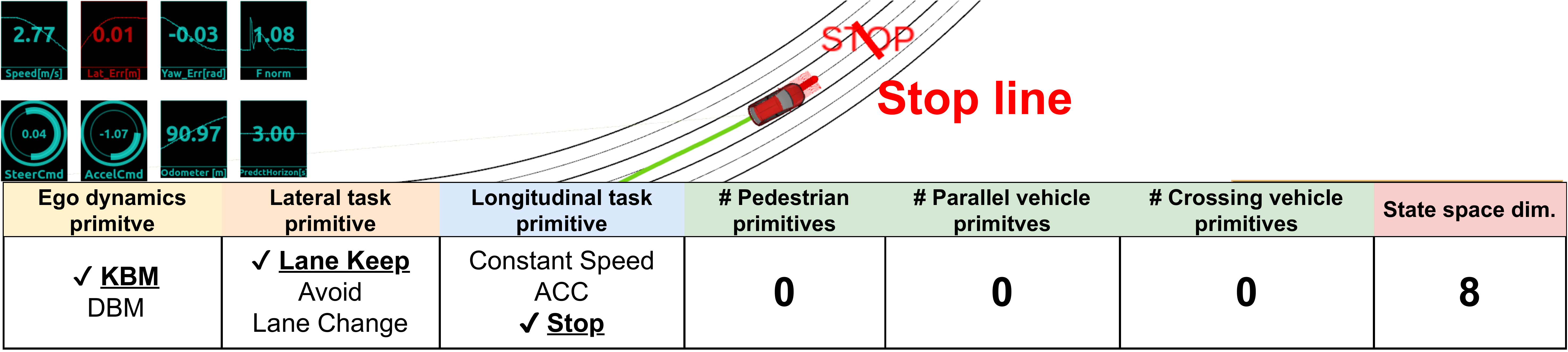}
    \subcaption{Smoothly stopping at a stop line}
    \label{fig:shared_space_2}
  \end{minipage}
  \caption{Result of scenario SD. The ego vehicle yields the way to crossing pedestrians (a) and stops at the stop line (b). Because our proposed system switches vehicle models according to the speed range, the ego vehicle can drive stably at slow and high speeds.}
  \label{fig:shared_space}
   % \vspace{-3mm}
\end{figure}

We can also observe three scenarios in scenario SD in Fig. \ref{fig:shared_space}.
The KBM primitive kept the ego car stable despite low driving speeds and large steering angles.
Consequently, the ego vehicle slowed for pedestrian crossing and then smoothly stopped at the stop line.

\subsubsection{Overview of the simulation results}

The results of 100 runs for each scenario are summarized in Table \ref{tab:result}.
The number of generated MPCs significantly exceeded the number of MPC primitives used.
This result shows that our proposed framework represents and realizes various driving tasks by selecting the appropriate MPC primitives from the pool, instead of manually designing the task-specific MPC for each driving task.

Next, the drivability of the proposed framework was discussed.
A successful trial is defined as the completion of the driving task without any collision or departure from lanes. 
Out of the 100 trials, our proposed method succeeded 87 times in AL, 92 times in OA, 91 times in TI, and 84 times in SD.
While in many cases, it succeeded, it failed in some cases.
The main reason for these failures was the failure in optimization when a new MPC primitive was added.
The proposed framework cannot guarantee the feasibility of the generated MPCs when the optimization problem is updated, because a new MPC primitive may have constraints that cannot be satisfied at the time instance.
For example, imagine a case in which the parallel vehicle travels right next to the ego vehicle in the target lane when a lane change primitive is added. 
In this case, the safety constraint to maintain a distance from the other vehicle was already violated.
We plan to solve this problem by applying another method that maintains the feasibility of switching optimization problems \cite{honda_impc}.

\begin{table}[t]
\centering
% \vspace{-2mm}
\caption{Simulation Result}
\begin{tabular}{ccccc}
\hline
Scenario              & AL       & OA       & TI        & SD       \\ \hline
No. of used MPC primitives & 6        & 6        & 5         & 6        \\
No. of set of generated MPCs & 26       & 30       & 11        & 16       \\
Success rate [\%]     & 87       & 92       & 91        & 84       \\ \hline
\end{tabular}
\label{tab:result}
 % \vspace{-3mm}
\end{table}

\subsection{Discussion on computation time}

\begin{figure}[t]
  \centering
  \includegraphics[width=\linewidth]{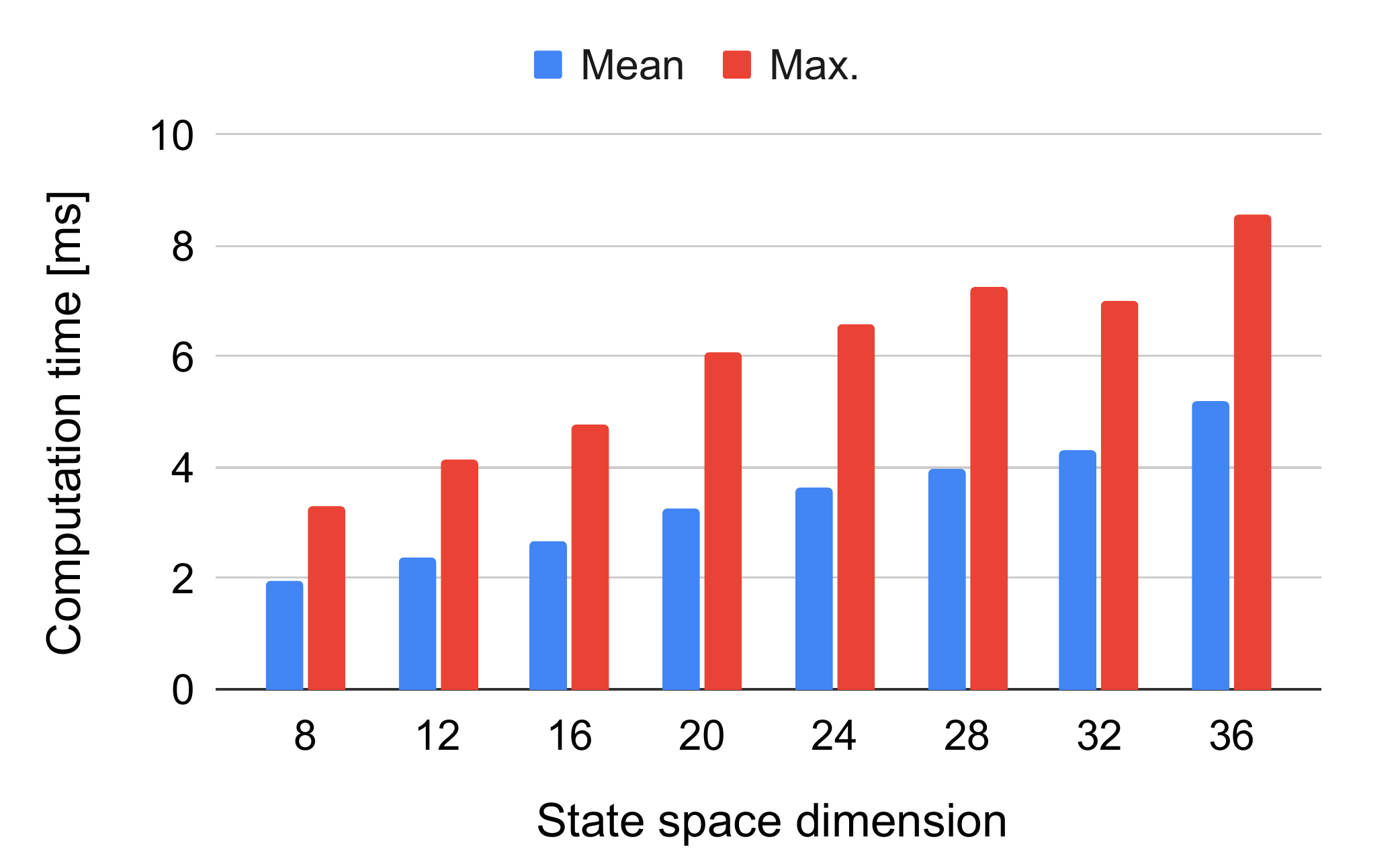}
  % \vspace{-2mm}
	\caption{ Mean and maximum computation time. The times for generating and solving MPCs over all simulations are shown corresponding to the state space dimensions. Although the computation time increases as the state space size, the computation finished within 10ms.}
  % \vspace{-5mm}
  \label{fig:computation_time}
\end{figure}

Our proposed method includes the states of the ego vehicle and other agents in the MPC state spaces. 
Therefore, the MPC can plan a trajectory considering the predicted behavior of the surrounding others.
However, the larger the state space, the greater the computational load for optimization.
Owing to the superior computational performance of C/GMRES, the real-time performance was maintained even when the state space became large.
Figure \ref{fig:computation_time} shows the dimensions of the state space and the mean and maximum computation times for all simulations.
A desktop PC with an Intel Core i9-10850K CPU was used to perform the simulations.
Although the computation time increased with the state-space dimension, both building MPCs and solving the optimization problems were performed in real time within 10 ms to update the control input.
Note that the number of finding variables, including the state prediction and the series of control inputs, is up to approximately 10,000 as the number of prediction steps $N=300$.

\section{CONCLUSION}
This study presents a vehicle motion planning and control framework for automatically generating MPCs according to various driving situations.
The key idea of the proposed framework is to define MPC primitives as the small components of MPCs to achieve each control requirement, and compose them to express various driving tasks in real time.
The MPC primitive is a parametrized optimization problem.
A binary operation to combine a set of MPC primitives is defined and applied in every control cycle based on the measured driving situation.
This scheme enables the representation of various behaviors by combining fewer elements instead of manually designing complex MPCs for many desired tasks commonly used in a simple conventional MPC switching framework.

\bibliographystyle{IEEEtran}
\bibliography{IEEEabrv, reference}

\end{document}